\documentclass[runningheads]{llncs}
\usepackage{bm}
\usepackage{amsmath}
\usepackage{threeparttable}
\usepackage{multirow}
\usepackage{graphicx}
\usepackage{url}
\graphicspath{{img/}}
\DeclareGraphicsExtensions{.png,.pdf}
\usepackage[caption=false,font=normalsize,labelfont=sf,textfont=sf]{subfig}
\begin{document}
\title{A Unified Mammogram Analysis Method via \\Hybrid Deep Supervision}
%
%
\author{Rongzhao Zhang, Han Zhang, Albert C.S. Chung
}
%
%
\institute{The Hong Kong University of Science and Technology
}
\maketitle              
\begin{abstract}
Automatic mammogram classification and mass segmentation play a critical role in a computer-aided mammogram screening system. In this work, we present a unified mammogram analysis framework for both whole-mammogram classification and segmentation. Our model is designed based on a deep U-Net with residual connections, and equipped with the novel hybrid deep supervision (HDS) scheme for end-to-end multi-task learning. As an extension of deep supervision (DS), HDS not only can force the model to learn more discriminative features like DS, but also seamlessly integrates segmentation and classification tasks into one model, thus the model can benefit from both pixel-wise and image-wise supervisions. We extensively validate the proposed method on the widely-used INbreast dataset. Ablation study corroborates that pixel-wise and image-wise supervisions are mutually beneficial, evidencing the efficacy of HDS. The results of 5-fold cross validation indicate that our unified model matches state-of-the-art performance on both mammogram segmentation and classification tasks, which achieves an average segmentation Dice similarity coefficient (DSC) of 0.85 and a classification accuracy of 0.89. The code is available at \url{https://github.com/angrypudding/hybrid-ds}.

\end{abstract}
\section{Introduction}
Breast cancer is one of the top causes of cancer death in women. In 2017, it is estimated that there are 252,710 new diagnoses of invasive breast cancer among women in the United States, and approximately 40,610 women are expected to die from the disease \cite{desantis2017breast}. The detection of breast cancer in its early stage by mammography allows patients to get better treatments, and thus can effectively lower the mortality rate. Currently, mammogram screening is still based on experts’ reading, but this process is laborious and prone to error.

Computer-aided diagnosis (CADx) system is a potential solution to facilitate mammogram screening, and the research on automatic (or semi-automatic) mammogram analysis has been a focus in medical vision field. Given the fact that a mass only occupies a small region (typically ~2\%) of a whole mammogram (i.e. the “needle in a haystack” problem \cite{lotter2017multi}), it is very hard to identify a mass from the whole image without introducing a large number of false positives. Therefore, traditionally, both hand-crafted feature based methods \cite{buciu2011directional,pratiwi2015mammograms} and deep learning models \cite{geras2017high,kooi2017large} require manually extracted regions of interest (ROIs), which, however, affects their usefulness in clinical practice. Recently, Dhungel et al. \cite{dhungel2017deep} proposed a sophisticated framework integrating mass detection, segmentation and classification modules to do whole-image classification, which achieved state-of-the-art performance with minimal manual intervention (manually rejecting false positives after detection). Besides, Lotter et al. \cite{lotter2017multi} proposed a 2-stage curriculum learning method to cope with the classification of whole mammograms, and Zhu et al. \cite{zhu2017deep} developed a sparse multi-instance learning (MIL) scheme to facilitate the end-to-end training of convolution neural networks (CNNs) for whole-image classification. Nevertheless, these methods either require manual intervention and multi-stage training, or only focus on the classification problem, while the accurate location and size of masses also play a critical role in a CADx system.

In this paper, we propose a CNN-based model with Hybrid Deep Supervision (Hybrid DS, HDS) to perform whole-mammogram classification and mass segmentation simultaneously. This model is based on a very deep U-Net \cite{ronneberger2015u} with residual connections \cite{he2016deep} (U-ResNet) which has 45 convolutional layers in the main stream. To facilitate the multi-task training of the deep network and boost its performance, we extend deep supervision (DS) \cite{lee2015deeply} to Hybrid DS by introducing multi-task supervision into each auxiliary classifier in DS, and apply this scheme to the U-ResNet model. To evaluate the proposed method, we performed extensive experiments on a publicly available full-field digital mammographic (FFDM) dataset, i.e. INbreast \cite{moreira2012inbreast}. The results show that our model achieves state-of-the-art performance in both classification and segmentation metrics, and ablation studies are performed to demonstrate the efficacy of HDS scheme.

\section{Method}
\begin{figure}
\includegraphics[width=\textwidth]{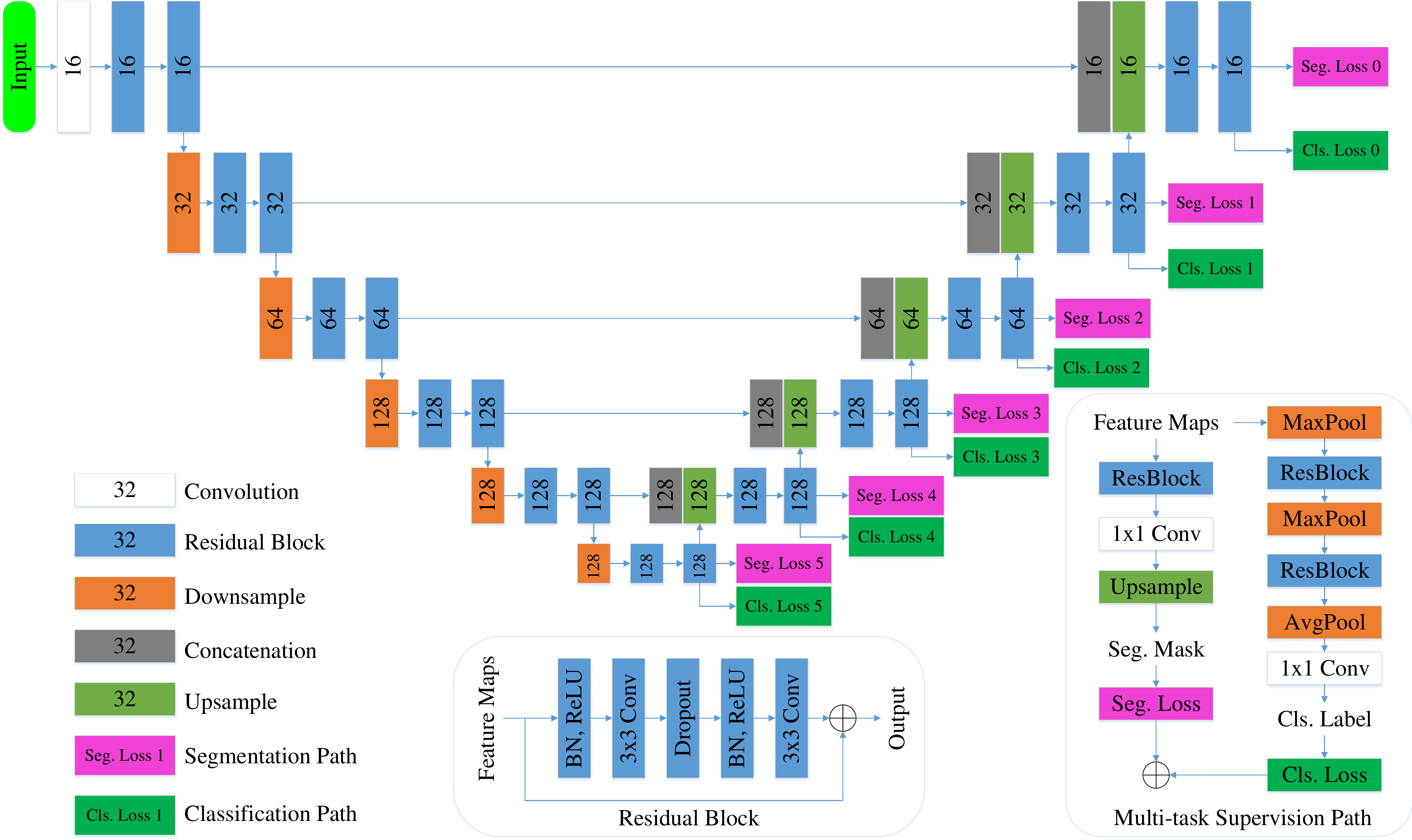} %
\caption{Architecture of the U-ResNet model with Hybrid DS. Best viewed in color.} \label{fig_archi}
\end{figure}

\subsection{Motivation}
Due to the very small size of masses, directly training deep CNN models for whole mammogram classification can lead to a severe overfitting problem, where the powerful model may easily memorize the patterns presented in the background area rather than learn the feature of masses, leading to poor generalization performance. To deal with this problem, we propose to employ both image-wise and pixel-wise labels to supervise the training process. The underlying assumption for this multi-task scheme is two-fold. First, since classification (whether there exist any masses in a mammogram) and segmentation (whether each pixel belongs to a mass) are highly correlated tasks, the features learned in one task should also be useful in the other; second, multi-task learning itself can serve as a regularization method as it prevents the training process from biasing towards either task. Therefore, we propose a multi-task CNN model trained with Hybrid DS to attack the whole-mammogram classification and segmentation problems.

\subsection{Hybrid Deep Supervision}
Similar to DS, Hybrid DS directly supervises intermediate layers via auxiliary classifiers to force the model to learn more discriminative features in early layers. Meanwhile, HDS extends DS by introducing multi-task classifiers into each supervision level. Formally, the optimization objective of HDS is defined as:
\begin{equation}\label{hds1}
  \begin{aligned}
  \mathcal{L}(\bm X;\bm W,\widehat{\bm w}) = &\mathcal{L}^{(\rm{seg})}(\bm X;\bm W,\widehat{\bm w}^{(\rm{seg})}) + \alpha\mathcal{L}^{(\rm{cls})}(\bm X;\bm W,\widehat{\bm w}^{(\rm{cls})}) \\
  &+ \lambda(\|\bm W\|_2 + \|\widehat{\bm w}^{(\rm{seg})}\|_2 + \|\widehat{\bm w}^{(\rm{cls})}\|_2)
  \end{aligned}
\end{equation}
where $\bm X$ denotes the training dataset, $\bm W$ is the trainable parameters of the main stream, i.e. the U-ResNet model without multi-task supervision pathways, $\widehat{\bm w}^{(\rm{seg})}$ and $\widehat{\bm w}^{(\rm{cls})}$ respectively denote the parameters of the segmentation and classification parts of the multi-task paths, and $\alpha$ is a constant that controls the relative importance of the classification loss. The third term to the right is a regularization term weighted by a hyper-parameter $\lambda$, and $\|\cdot\|_2$ denotes L2-norm. The segmentation loss $\mathcal{L}^{(\rm{seg})}$ and classification loss $\mathcal{L}^{(\rm{cls})}$ are defined as follow:
\begin{align}\label{hds2}
  \mathcal{L}^{(\rm{seg})}(\bm X;\bm W,\widehat{\bm w}^{(\rm{seg})}) &= \sum_{d\in \mathcal{D}}\eta_d\mathcal{J}_d^{\rm{(seg)}}(\bm X; \bm W_d, \widehat{\bm w}^{\rm{(seg)}}_d) \\
  \mathcal{L}^{(\rm{cls})}(\bm X;\bm W,\widehat{\bm w}^{(\rm{cls})}) &= \sum_{d\in \mathcal{D}}\eta_d\mathcal{J}_d^{\rm{(cls)}}(\bm X; \bm W_d, \widehat{\bm w}^{\rm{(cls)}}_d)
\end{align}
where $\bm W_d$ denotes the parameter in the first $d$ layers of the main stream, $\widehat{\bm w}^{\rm{(seg)}}_d$ and $\widehat{\bm w}^{\rm{(cls)}}_d$ are respectively the weights in the segmentation and the classification parts of the multi-task path associated with the $d$-th layer, $\eta_d$ is the weights of the corresponding loss level, and $\mathcal{D}$ is a set that contains the indices of the layers directly connected to multi-task supervision paths. $\mathcal{J}_d^{\rm{(seg)}}$ is a segmentation cross-entropy cost function that returns the average loss across all pixel locations. $\mathcal{J}_d^{\rm{(cls)}}$ is basically a cross-entropy cost function, which is made compatible with MIL scheme, and its definition will be detailed in the following section.

\subsection{Deep Multi-Instance Learning}
Conventionally, CNN models used in classification tasks includes at least 1 fully connected (FC) layers, which can only take fixed-size inputs. However, segmentation models, e.g. fully convolutional networks (FCNs), are usually trained on cropped image patches and tested on whole images, where the input size may vary. To integrate the two tasks into one unified framework, we convert the classification part of the model into a FCN manner. Thus, the classification part may take inputs of different sizes like a segmentation model, but its output also turns into a 2D probabilistic map, no longer a single value. If we map the pixels in such a 2D map back to nonoverlapping patches in the input image, the whole input image can then be regarded as a bag of patches (instances), thus the mammogram classification can be treated as a standard MIL problem. In this case, denoting the pixel values in a 2D probabilistic map as $r_{i,j}$, the mass probability of the input image $\bm I$ is then $p(y=1|\bm I) = \max_{i,j}\{r_{i,j}\}$. Following the practice of Zhu et al. in \cite{zhu2017deep}, we define the classification cost for an input image $\bm I$ as below:
\begin{equation}\label{eq_mil}
  \mathcal{J}^{\rm {(cls)}}(\bm I,y_I;\bm W, \bm w^{\rm{(cls)}}) = -\log p(y=y_I|\bm I) + \mu\sum_{i,j}r_{i,j}
\end{equation}
where $y_I$ is the true label of image $\bm I$, and $r_{i,j}$ is the pixel value in the 2D probabilistic map. Since masses are sparse in mammograms, the summation of $r_{i,j}$ should be small. Therefore, a sparsity term (the second term to the right) is added to the cost function, which is weighted by $\mu$.

\subsection{Network Architecture}
The architecture of the proposed neural network model is illustrated in Fig. \ref{fig_archi}. The model is basically a deep U-ResNet with 45 $3\times3$ convolutional layers (1 convolution and 22 residual blocks), and multi-task supervision pathways are inserted into each scale level for Hybrid DS. We use max pooling in downsampling modules (except for the last downsampling layer in the classification part of each multi-task path, which employs average pooling), and bi-linearly upsample feature maps in upsampling layers. For those transition modules (i.e. downsample, upsample and concatenation), if the input and output channel dimensions are different, $1\times1$ convolutions are inserted before the operation to change the channel dimension. All max pooling layers have a stride of 2, and the stride of average pooling layers ranges from $2^0$ to $2^5$ to ensure a total downsampling factor of $2^7$ for the output of each classification path (so the size of the output probabilistic map is $4\times3$ in training and $8\times4$ in testing). Similarly, All upsampling layers except for the ones in multi-task paths have a stride of 2, and those in multi-task paths range from $2^0$ to $2^5$ to ensure the output mask have the same size as the input image. Besides, Dropout \cite{srivastava2014dropout} layers of rate 0.2 (for residual blocks with less than 128 channels) or 0.5 (for others) are inserted into each residual block.

\section{Experiments and Results}
\subsubsection{Dataset}
The proposed method was evaluated on a publicly available FFDM dataset, i.e. INbreast \cite{moreira2012inbreast}. Among the 410 mammograms in INbreast dataset, 107 contain one or more masses, and totally contain 116 benign or malignant masses. In pre-processing, we removed the left or right blank regions by thresholding, resized the mammograms to $1024\times512$, and then normalized each image to zero-mean and unit-std according to the statistics of training sets. During training, the whole image was randomly flipped vertically or horizontally, and patches of size $512\times384$ were randomly sampled from it with 50\% chance centered on a positive (mass) pixel. The classification label of a cropped patch was set to 1 if the patch contained any pixel from masses, and 0 otherwise. In our experiment, the whole dataset was uniformly divided into 5 folds (82 images per fold), and we used three of them for training, one for validation and one for testing. We first performed ablation study on one data split to demonstrate the efficacy of the proposed method, and then ran a 5-fold cross validation for a fairer comparison with existing methods.

\subsubsection{Implementation Details}
The proposed method was implemented with PyTorch v0.4.0 on a PC with one NVIDIA Titan Xp GPU. Stochastic gradient descent (SGD) method with a momentum of 0.9 was used to optimize the model, with an initial learning rate of 0.01 and decayed by 0.3 after 1000, 1800, 2400 and 2410 epochs. In all experiments, the model was trained for 2800 epochs, which took about 12.5 hours and was long enough for each configuration to converge. The model parameters were initialized by Kaiming method \cite{he2015delving}. Other hyperparameters were set as follows: classification loss weight $\alpha=0.03$ (such that the segmentation and classification losses of each mini-batch were comparable in magnitude), weight decay $\lambda=0.0005$, sparsity weight $\mu=10^{-6}$, and the weights of different supervision levels $(\eta_0,\eta_1,\eta_2,\eta_3,\eta_4,\eta_5)=(1.0, 1.5, 2.0, 2.5, 3.0, 3.5)$, where $\eta_1$ to $\eta_5$ were gradually decayed to very small values (i.e. 0.005$\eta_i$) during training. The losses stemmed from inner layers of the U-Net were initially weighted higher to force the these layers to learn meaningful features, otherwise they tended to be ignored due to the difficulty in learning from low-resolution feature maps.

\subsubsection{Metrics}
We employed dice similarity coefficient (DSC), sensitivity (SE) and false positives per image (FPI) to evaluate the segmentation results. For classification, accuracy (ACC), area under ROC curve (AUC), $F_1$ score, precision (Prec) and recall (Recl) were reported.

\begin{table}[]
\centering
\caption{Ablation Study}
\label{tab_ablation}
\begin{threeparttable}
\setlength{\tabcolsep}{2mm}{
\begin{tabular}{lllllllll}
\hline
\multirow{2}{*}{Model} & \multicolumn{3}{c|}{Segmentation}                  & \multicolumn{5}{c}{Classification}                                                                                                 \\ \cline{2-9}
                       & \multicolumn{1}{c}{DSC} & \multicolumn{1}{c}{SE} & \multicolumn{1}{c|}{FPI} & \multicolumn{1}{c}{ACC} & \multicolumn{1}{c}{AUC} & \multicolumn{1}{c}{$F_1$} & \multicolumn{1}{c}{Recl} & \multicolumn{1}{c}{Prec} \\ \hline
Multi-task Only        & 0.787                   & 0.882       &0.293         & 0.829                   & 0.866                   & 0.682                     & 0.714                    & 0.652                    \\
Cls + DS               & N/A                     & N/A         &N/A            & 0.878                   & 0.853                   & 0.722                     & 0.619                    & 0.867                    \\
Seg + DS$^*$          & 0.802                   & \textbf{0.910}     &0.183  & 0.842                   & \textbf{0.890}          & 0.723                     & \textbf{0.810}           & 0.654                    \\
Hybird DS              & \textbf{0.848}          & 0.907         &0.110      & \textbf{0.915}          & 0.887                   & \textbf{0.821}            & 0.762                    & \textbf{0.889}           \\ \hline
\end{tabular}}
\begin{tablenotes}
    \item $^*$Classification results were retrieved from segmentation masks by assigning the largest activation across the output probabilistic map to the whole image.
\end{tablenotes}
\end{threeparttable}
\end{table}

\subsubsection{Ablation Study}
To investigate the efficacy of the proposed Hybrid DS scheme, a series of experiments were conducted on one data split. From Table \ref{tab_ablation}, it can be observed that the Hybrid DS model achieves the best performance on several important metrics (e.g. DSC, FPI, ACC, $F_1$, etc.), and also has high scores on others. HDS outperforms the baseline multi-task model by a large margin (0.848 vs 0.787 in DSC, 0.915 vs 0.829 in ACC), indicating that directly supervising intermediate layers is necessary for training such a deep model. Thanks to the sparse MIL \cite{zhu2017deep} and DS schemes, the Cls+DS model performed well in classification, having an accuracy of 0.878. Meanwhile, HDS achieves even higher classification performance (e.g. ACC: 0.915) than Cls+DS, which evidences the benefit of employing extra pixel-wise supervision. Compared to Seg+DS, HDS achieves better DSC (0.848), accuracy (0.915), $F_1$ score (0.821) and precision (0.889), which we attribute to the extra image-wise supervision. Since image-wise supervision can force the network to look wider and to learn features based on the whole image (or at least a larger area), the network becomes less sensitive to local patterns that mimic masses and more robust in rejecting false positives, as has been validated by the much higher precision of HDS (0.889) than Seg+DS (0.654). Altogether, these experiments suggest that the proposed Hybrid DS scheme is a promising approach to improve deep model's performance on the mammogram analysis problem.

\begin{table}[b]
\centering
\caption{Comparison with State-of-the-art Methods}
\label{tab_comparison}
\begin{threeparttable}
\begin{tabular}{lllllllll}
\hline
\multirow{2}{*}{Model} & \multicolumn{3}{c|}{Segmentation}                  & \multicolumn{5}{c}{Classification}                                                                                                 \\ \cline{2-9}
                       & \multicolumn{1}{c}{DSC} & \multicolumn{1}{c}{SE} & \multicolumn{1}{c|}{FPI} & \multicolumn{1}{c}{ACC} & \multicolumn{1}{c}{AUC} & \multicolumn{1}{c}{$F_1$} & \multicolumn{1}{c}{Recl} & \multicolumn{1}{c}{Prec} \\ \hline
D. \cite{dhungel2017deep} & 0.85$^*$             & N/A     & 1.00$^\#$          & 0.91$\pm$0.02           & 0.76$\pm$0.23           & N/A                       & N/A                      & N/A                      \\
Z. \cite{zhu2017deep}  & N/A                     & N/A       & N/A              & 0.90$\pm$0.02           & 0.89$\pm$0.04           & N/A                       & N/A                      & N/A                      \\
Ours                   & 0.85$\pm$0.01            & 0.88$\pm$0.02   & 0.08$\pm$0.02 & 0.89$\pm$0.02            & 0.85$\pm$0.02            & 0.77$\pm$0.04              & 0.69$\pm$0.04             & 0.87$\pm$0.06             \\ \hline
\end{tabular}
\begin{tablenotes}
    \item $^*$Calculated on correctly detected masses.
    \item $^\#$These detection false positives were manually rejected before further processing.
\end{tablenotes}
\end{threeparttable}
\end{table}

\subsubsection{Comparison with Existing Methods}
To compare the proposed model with other mammogram analysis methods, we used 5-fold cross validation to evaluate it on the whole INbreast dataset. As shown in Table \ref{tab_comparison}, our model matches the current state-of-the-art performance on both mass segmentation and classification tasks, achieving a high average segmentation DSC of 0.85 and a classification accuracy of 0.89. Besides, our method is fully automatic and easy to deploy, which takes whole mammograms as input and then outputs segmentation masks and image-wise labels simultaneously. In contrast, the method by Dhungel et al. \cite{dhungel2017deep} still requires manual intervention to reject false positives after mass detection, and the method by Zhu et al. \cite{zhu2017deep} can only give a very rough location of identified masses. Qualitative segmentation results of our method on several typical testing images have been illustrated in Fig. \ref{fig_quali}.

\newcommand{\picwidthVII}{0.7in}
\begin{figure*}[t]
	\captionsetup[subfigure]{labelformat=empty}
	\centering
    \subfloat{\includegraphics[width=\picwidthVII]{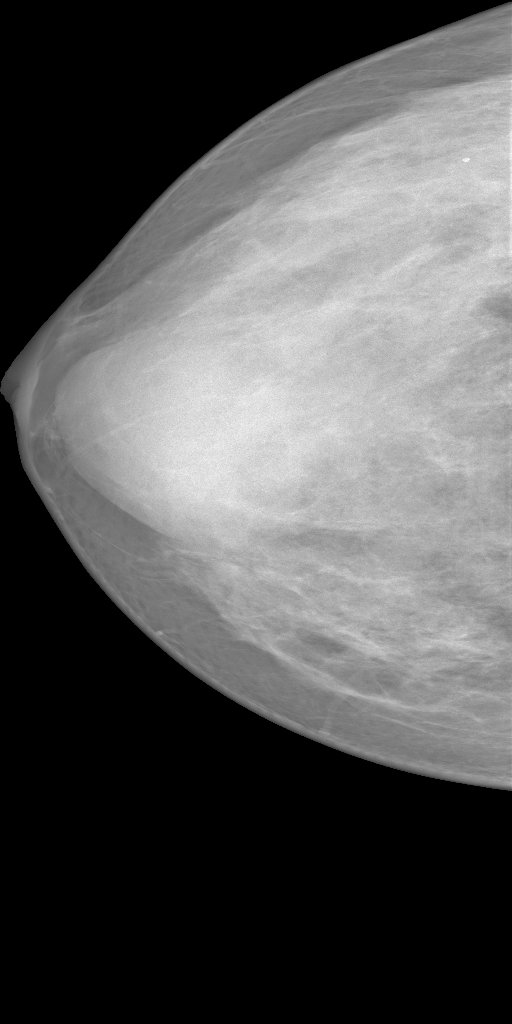}}
	\hspace{0in}	
	\subfloat{\includegraphics[width=\picwidthVII]{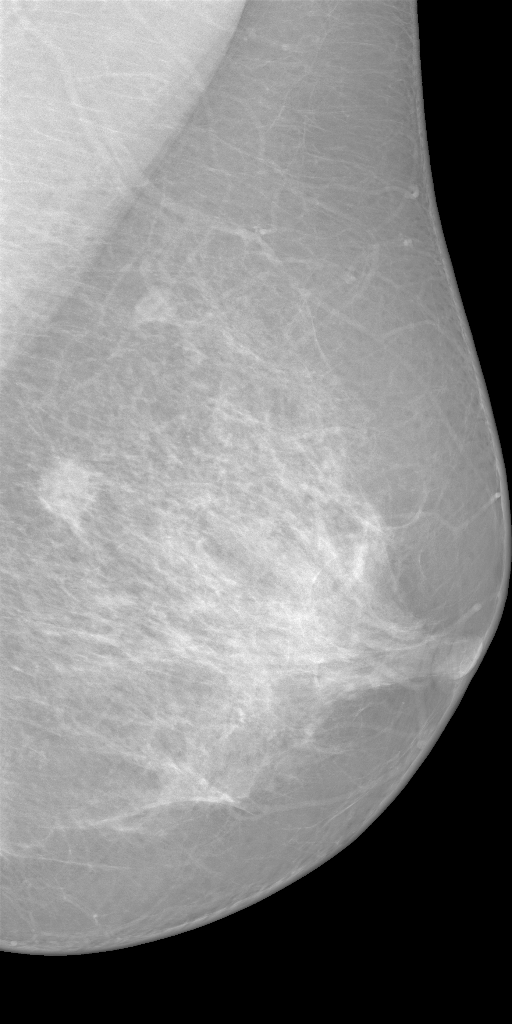}}
	\hspace{0in}	
	\subfloat{\includegraphics[width=\picwidthVII]{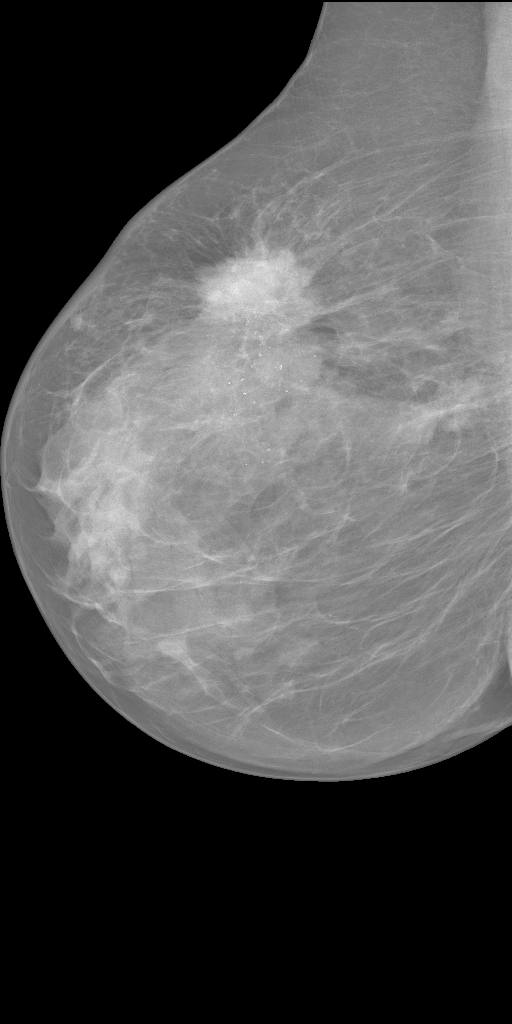}}
	\hspace{0in}	
	\subfloat{\includegraphics[width=\picwidthVII]{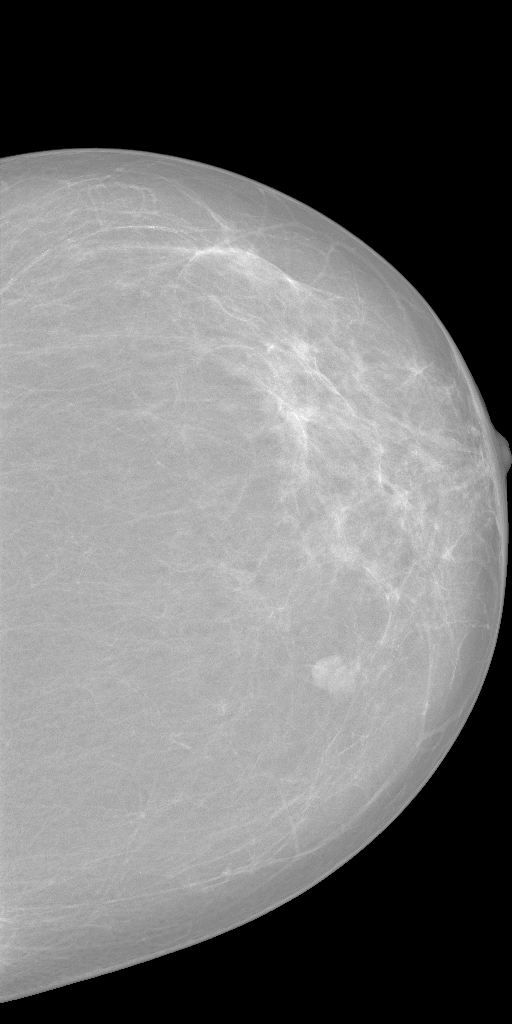}}
	\hspace{0in}	
	\subfloat{\includegraphics[width=\picwidthVII]{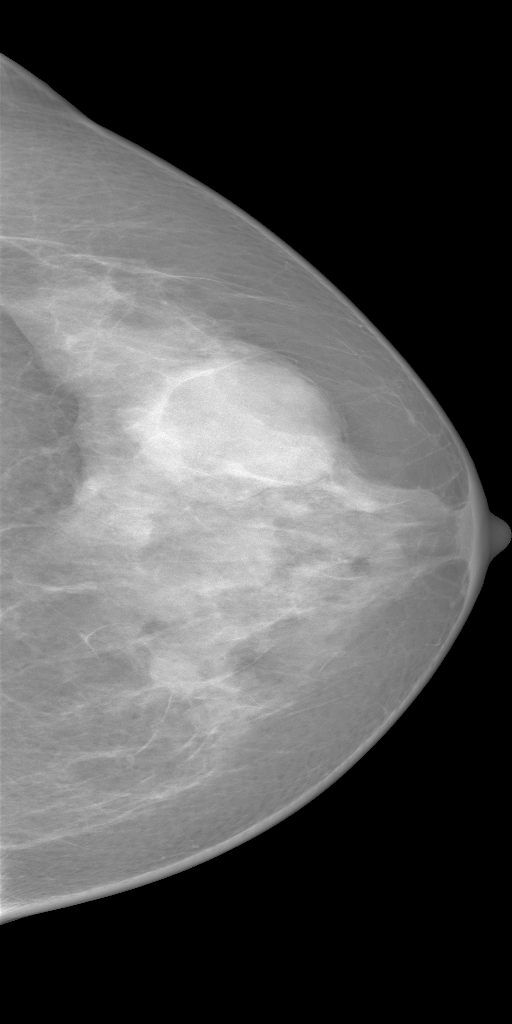}}
	\hspace{0in}	
	\subfloat{\includegraphics[width=\picwidthVII]{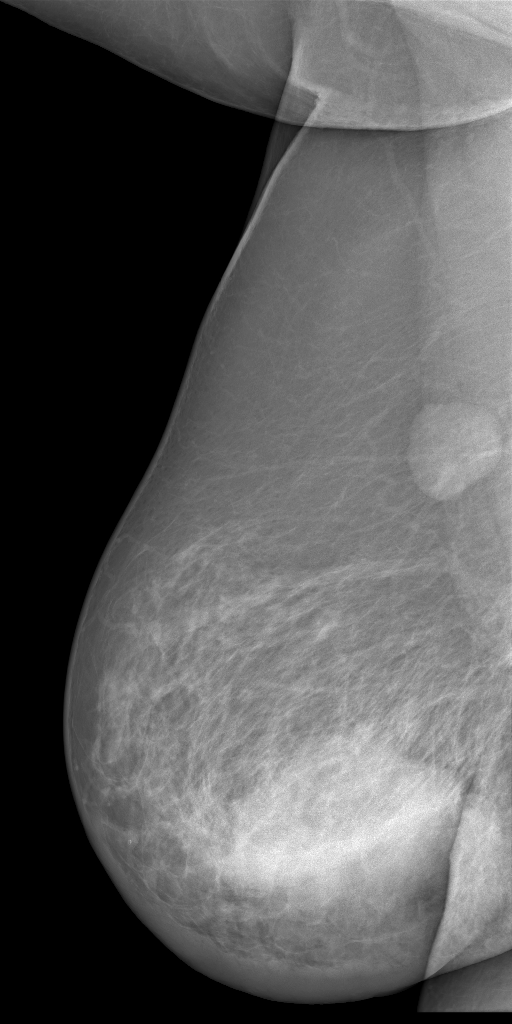}}

	\subfloat{\includegraphics[width=\picwidthVII]{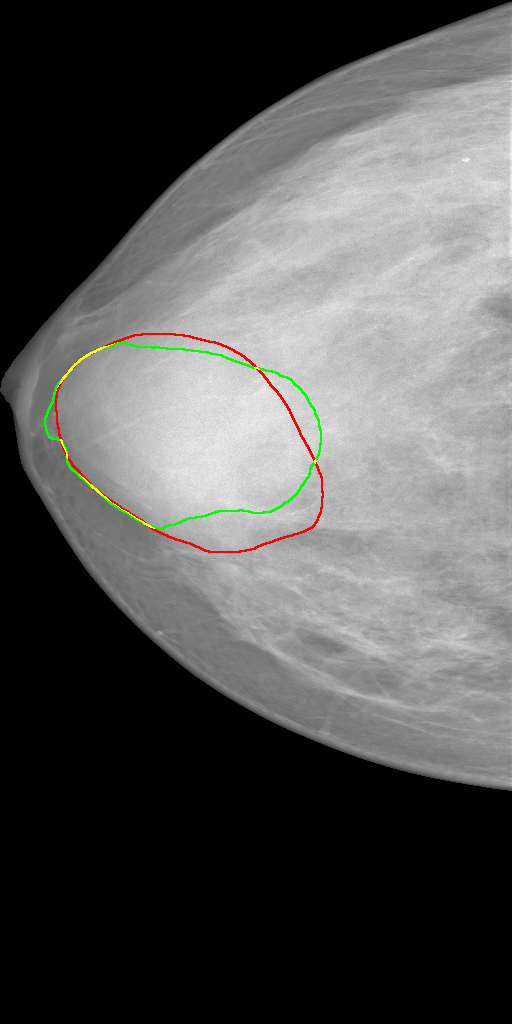}}
	\hspace{0in}	
	\subfloat{\includegraphics[width=\picwidthVII]{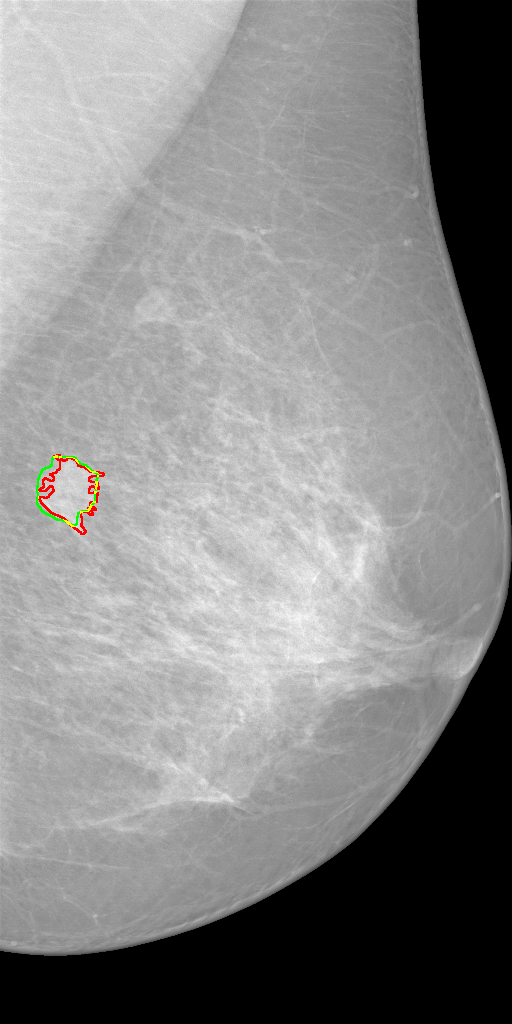}}
	\hspace{0in}	
	\subfloat{\includegraphics[width=\picwidthVII]{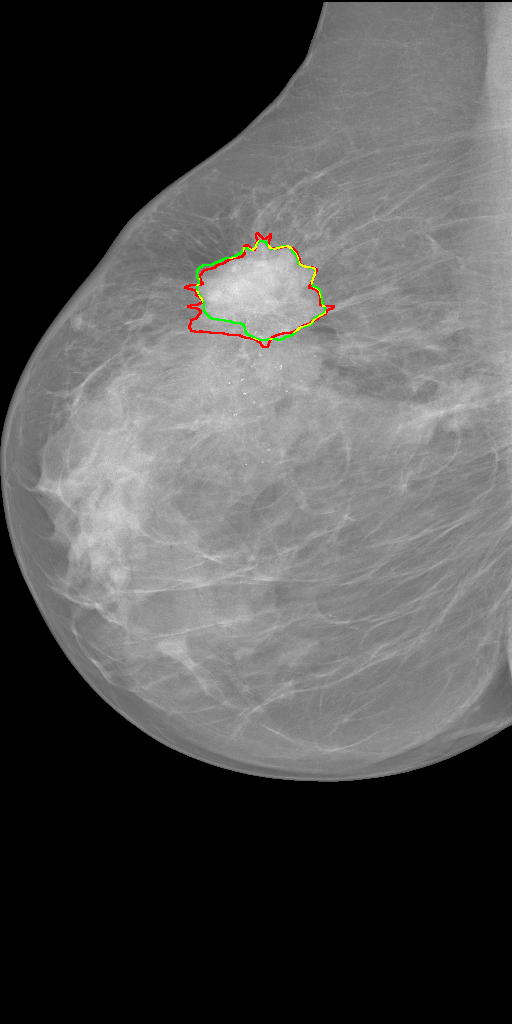}}
	\hspace{0in}	
	\subfloat{\includegraphics[width=\picwidthVII]{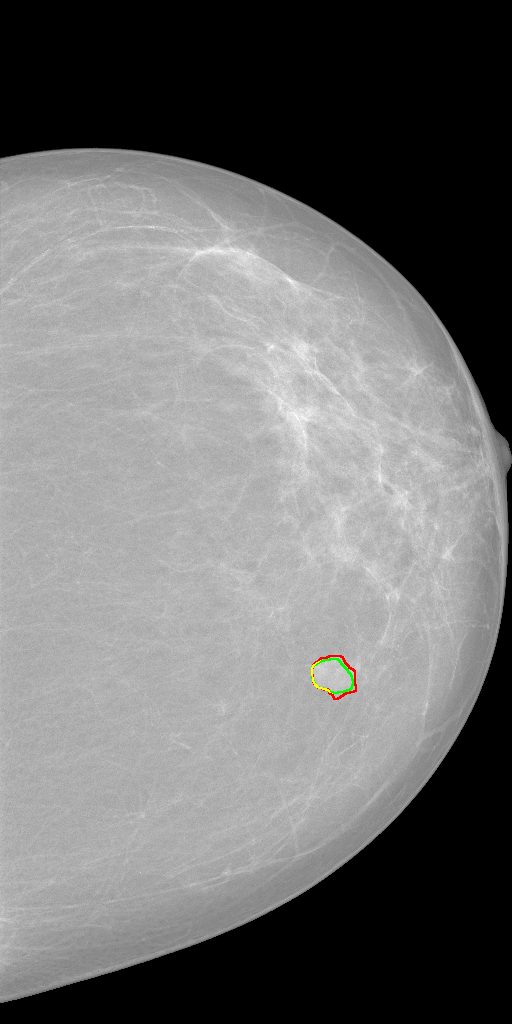}}
	\hspace{0in}	
	\subfloat{\includegraphics[width=\picwidthVII]{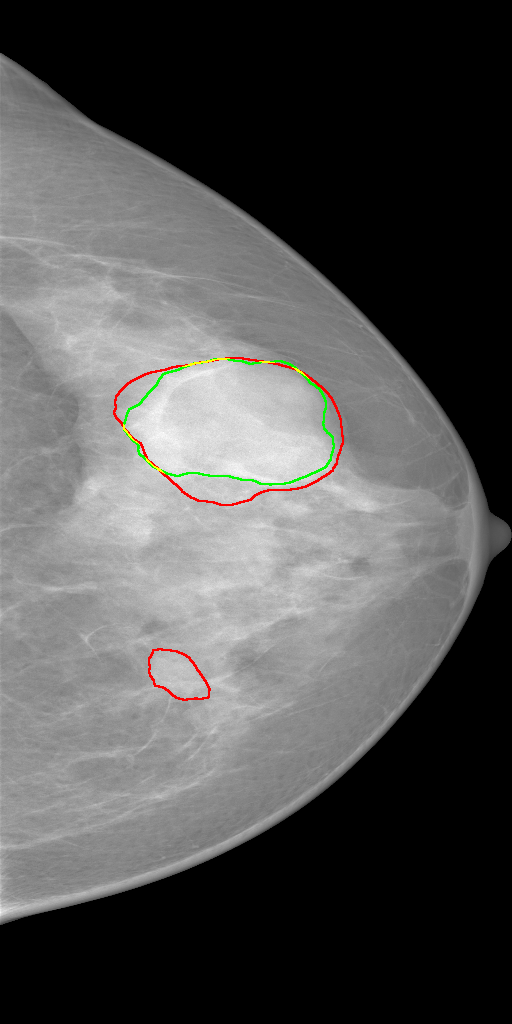}}
	\hspace{0in}	
	\subfloat{\includegraphics[width=\picwidthVII]{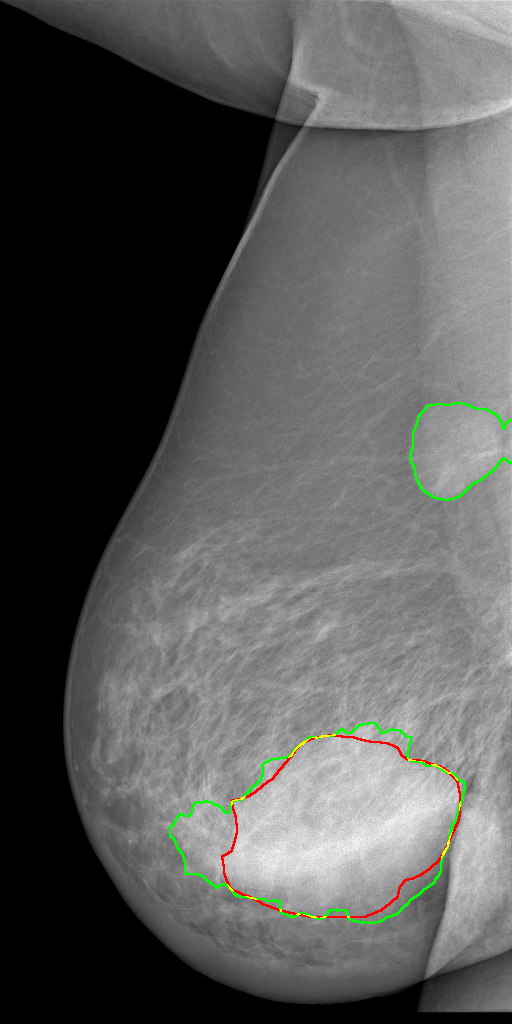}}
	\caption{Qualitative results. The first row is original mammograms. In the second row, red and green boundaries are ground truth delineation and automatic segmentation results, respectively. The fifth column contains a false negative lesion, and the last column has a false positive lesion. Best viewed in color.}
	\label{fig_quali}
\end{figure*}

\section{Conclusion}
In this paper, we have developed an end-to-end and unified framework for mammogram classification and segmentation. We seamlessly integrate the two tasks into one model by employing the novel Hybrid DS scheme, which not only inherits the merits of DS but also supports multi-task learning. With such a multi-task learning scheme, pixel-wise labels tell the model where to learn while image-wise labels force the network to make better use of contextual information. We conducted extensive experiments on the publicly available INbreast dataset, and the results show that our method matches the state-of-the-art performance on both segmentation and classification tasks. Ablation study demonstrates that pixel-wise and image-wise supervisions are mutually beneficial, and the proposed Hybrid DS can effectively boost the model's performance on both tasks. Besides, our unified framework is inherently general, which can be easily extended to other medical vision problems.

%
%
%
\bibliographystyle{splncs04}
\bibliography{bibtex}

\end{document}